%% file: acl2020.tex
\documentclass[11pt,a4paper]{article}
\pdfoutput=1
\usepackage[hyperref]{acl2020}
\usepackage{times}
\usepackage{latexsym}
\usepackage{framed}
\usepackage{amsmath} 
\usepackage{graphicx}
\usepackage{dsfont}
\usepackage{subfigure}
\usepackage{pdfcomment}
\usepackage[official]{eurosym}
\usepackage{bbm}

\usepackage{enumitem}
\setlist{nolistsep}

\DeclareMathOperator{\RawCov}{RawCov}
\DeclareMathOperator{\NormCov}{NormCov}
\DeclareMathOperator{\SummaryScore}{SummaryScore}
\DeclareMathOperator{\Fluency}{Fluency}

\usepackage{microtype}

\aclfinalcopy % Uncomment this line for the final submission
\def\aclpaperid{2152} %  Enter the acl Paper ID here

\title{The Summary  Loop: \\ Learning to Write Abstractive Summaries Without Examples}

\author{Philippe Laban \\
  UC Berkeley \\
   \\
  \And
  Andrew Hsi \\
  Bloomberg\\
  \And
  John Canny \\
  UC Berkeley\\
  \And
  Marti A. Hearst \\
  UC Berkeley\thanks{~Author emails: \{phillab,canny,hearst\}@berkeley.edu, ahsil@bloomberg.net} \\
}

\date{}

\begin{document}
\maketitle
\begin{abstract}
This work presents a new approach to unsupervised abstractive summarization based on maximizing a combination of coverage and fluency for a given length constraint. It introduces a novel method that encourages the inclusion of key terms from the original document into the summary: key terms are masked out of the original document and must be filled in by a coverage model using the current generated summary. A novel unsupervised training procedure leverages this coverage model along with a fluency model to generate and score summaries.
When tested on popular news summarization datasets, the method outperforms previous unsupervised methods by more than 2 R-1 points, and approaches results of competitive supervised methods. Our model attains higher levels of abstraction with copied passages roughly two times shorter than prior work, and learns to compress and merge sentences without supervision.

\end{abstract}

\begin{figure}[th]
    \begin{framed}
        \small
        \textbf{Original Document}: \textbf{Chilean} \textbf{President} announced Wednesday that his country, which has been \textbf{paralyzed} by \textbf{protests} over the last two weeks, will no longer \textbf{host} two major international \textbf{summits}. [...] The President has now canceled the \textbf{hosting} of the \textbf{economic APEC forum} and \textbf{COP25 environmental summit}, which were both due to take place later this \textbf{year}. [...]
        
        \vspace{0.5em}
        \textbf{Masked Document}: \_\_\_\_ \_\_\_\_ announced Wednesday that his country, which has been \_\_\_\_ by \_\_\_\_ over the last two weeks, will no longer \_\_\_\_ two major international \_\_\_\_. [...] The \_\_\_\_ has now \_\_\_\_ the \_\_\_\_ of the \_\_\_\_ \_\_\_\_ \_\_\_\_ and \_\_\_\_ \_\_\_\_ \_\_\_\_, which were both due to take place later this \_\_\_\_. [...]
        
        \vspace{0.5em}
        \textbf{Summary Loop [10 word constraint]}: Pinera cancelled the APEC summit at Santiago.\\ \textit{Coverage Score: 0.22}
        
        \vspace{0.5em}
        \textbf{Summary Loop [24 word constraint]}: Pinera said Chileans have been canceled the hosting of the APEC summit, which was scheduled to take place in November.\\
        \textit{Coverage score: 0.33}
        
        \vspace{0.5em}
        \textbf{Summary Loop [45 word constraint]}: Sebastian Pinera announced Wednesday that his country will not hold the APEC summit, which was scheduled to take place in Santiago. Pinera said that Chileans had been paralyzed by protests over the last two weeks.\\
        \textit{Coverage score: 0.39}  
    \end{framed}
    \vspace{-1em}
    \caption{Motivating example. A document from CNN.com (keywords generated by masking procedure are bolded), the masked version of the article, and generated summaries by three Summary Loop models under different length constraints.}
    \label{fig:coverage_example}
\end{figure}

\section{Introduction}

\input{new_intro.tex}

\section{Related Work}

\textbf{Supervised Abstractive Summarization.} Sequence-to-sequence (seq2seq) \cite{sutskever2014sequence} models trained using teacher-forcing are the most common approach to abstractive summarization \cite{nallapati2016abstractive}. A common architecture is the Pointer-Generator \cite{see2017get}. Performance can further be improved by constraining the attention \cite{gehrmann2018bottom,gui2019attention,wang2019concept} and using pretrained Transformer-based language models \cite{lewis2019bart,chi2019cross,edunov2019pre}. Through architectural changes, the training procedure remains constant: using a large corpus of document-summary pairs, the model is trained to reproduce target summaries.

\textbf{Unsupervised Summarization.} Most unsupervised summarization work is extractive: sentences deemed relevant are pulled out of the original document and stitched into a summary, based on a heuristic for a sentence's relevance \cite{mihalcea2004textrank,barrios2015variations,west2019bottlesum}. \citet{nikolov2019abstractive}'s abstractive approach is partially unsupervised, not requiring parallel data, but only a group of documents and a group of summaries. In contrast, our work does not require any summaries, and is trained using only documents.
\citet{radford2019language} summarize documents using a language model (GPT2) in a Zero-shot learning setting. The model reads the document followed by a special token ``TL/DR'', and is tasked with continuing the document with a summary. Our work is an  extension of this work: we initialize our Summarizer model with a GPT2 and specialize it with a second unsupervised method.

\textbf{Summarization and Q\&A.} \citet{eyal2019question} and \citet{arumae2018reinforced} turn reference summaries into fill-in-the-blank (FIB) questions, either as an evaluation metric or to train an extractive summarization model. In this work, we directly generate FIB questions on the document being summarized, bypassing the need for a reference summary.

\citet{scialom2019answers}'s work stays closer to a Q\&A scenario, and uses a Question Generation module to generate actual questions about the document, answered by a  Squad-based \cite{rajpurkar2018know} model using the generated summary. We refrain from using actual questions because question generation remains a challenge, and it is unclear how many questions should be generated to assess the quality of a summary.

\textbf{RL in Summarization.} \citet{Paulus2018ADR} introduced Reinforcement Learning (RL) to neural summarization methods by optimizing for ROUGE scores, leading to unreadable summaries. Since then, Reinforcement Learning has been used to select sentences with high ROUGE potential \cite{chen2018fast}, or optimize modified versions of ROUGE that account for readability \cite{pasunuru2018multi}. In all cases, the reward being computed relies on a reference summary, making the methods supervised. We craft a reward that does not require a target summary allowing our training process to remain unsupervised.

\section{The Summary Loop}
\label{section:summary_loop}
For this work, the definition of a summary is:

\begin{quote}
    \centering
    ``A summary is a \underline{brief}, \underline{fluent} text that \underline{covers} the main points of an original document.''
\end{quote}

Brevity, fluency and coverage are the three pillars of a good summary. Under a length constraint, a good quality summary should contain as much information about the original document as possible while retaining fluent and coherent English. 

Subsection \ref{section:steps} lays out the steps in the Summary Loop. Subsections \ref{section:summodel}--\ref{section:fluencymodel}  specify how each component is represented by a neural network. Section~\ref{section:training} shows how to  train a summarizer model using this architecture in an unsupervised manner.\footnote{The code, model checkpoints and other resources are available at \url{https://github.com/CannyLab/summary_loop} .}

\subsection{Summary Loop Steps}
\label{section:steps}

\begin{figure}[!htbp]
    \centering
    \includegraphics[width=0.47\textwidth]{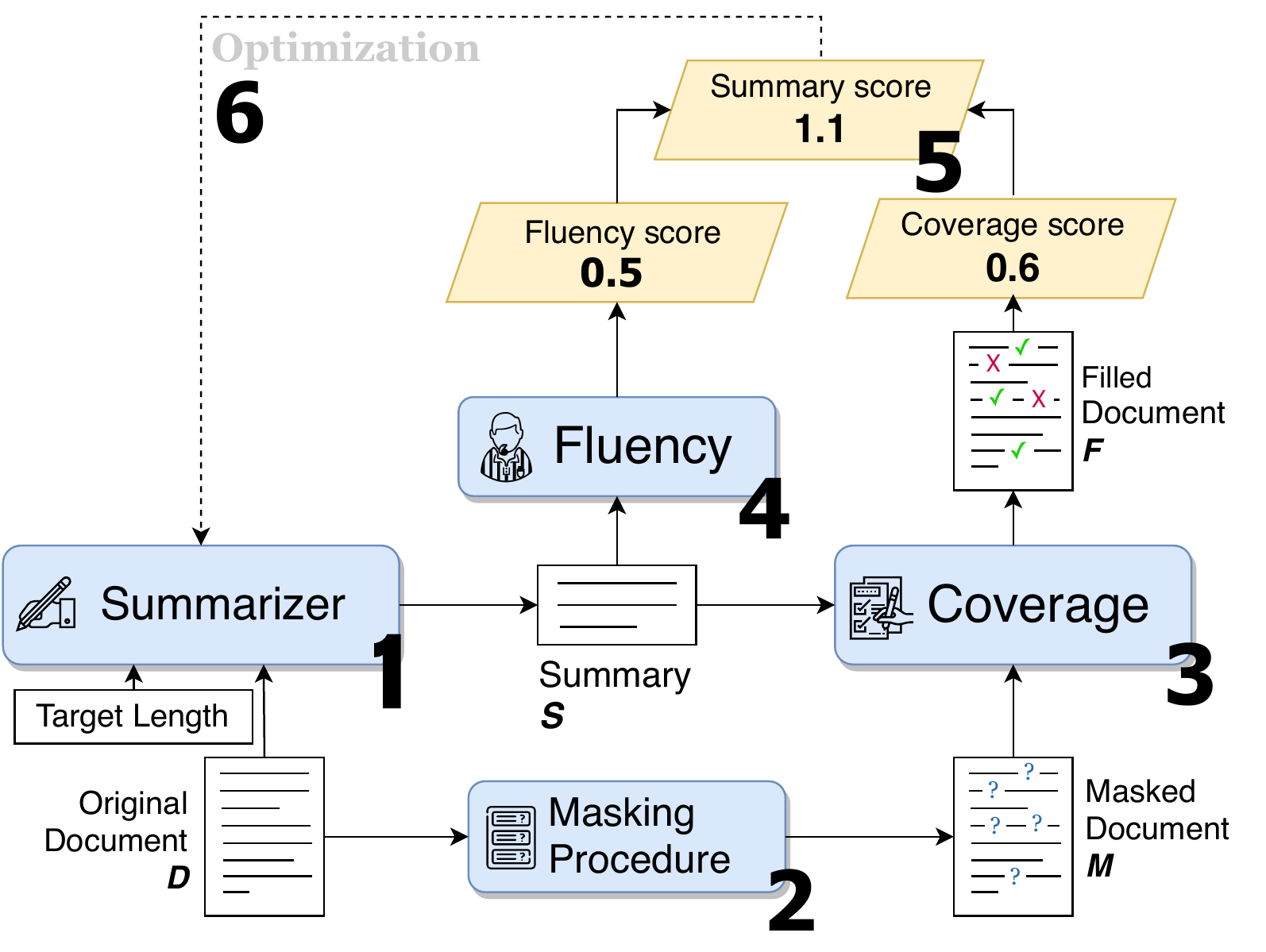}
    \caption{\textbf{The Summary Loop involves three neural models: Summarizer, Coverage and Fluency.} Given a document and a length constraint, the Summarizer writes a summary. Coverage receives the summary and a masked version of the document, and fills in each of the masks. Fluency assigns a writing quality score to the summary. The Summarizer model is trained, other models are pretrained and frozen.}
    \label{fig:loop_rules}
\end{figure}

Numbers in Figure~\ref{fig:loop_rules} correspond to the following steps:
\begin{enumerate}
	\item Summarizer receives a document D and length-constraint L, and produces a summary S fulfilling the length constraint.
	\item Using a Masking Procedure, D is modified into a masked document M, where important words have been replaced with blanks.
	\item Coverage receives S and M, and uses them to fill in each blank in M with a word, producing F. F and D are compared, and the resulting fill-in accuracy is called the Coverage Score.
	\item Fluency receives S, and gives a Fluency Score based on its assessment of the quality of the Summary's writing.
	\item The Fluency Score is added to the Coverage Score (as a weighed sum) into a Summary Score for the (D, S) pair.
	\item Reinforcement Learning is used to train the Summarizer to produce summaries with high Summary Score.
\end{enumerate}

The Summary Loop does not rely on the use of a target/reference/human-written summary, but only the summaries produced by the Summarizer model. The process can therefore be iterated upon without supervision from Summarization datasets.

% How does the new placement

\subsection{Summarization Model}
\label{section:summodel}

We use a Generative Transformer \cite{radford2019language} as the model architecture of the summarizer. We make this choice for two reasons. First, Generative Transformers can produce text one word at a time, allowing the system to produce abstractive summaries.
Second, we use the pretrained Generative Transformer to initialize the Summarizer.

Practically, the Summarizer first reads through the entire document, followed by a special \textit{START} token, signaling  summarization. The Summarizer produces a probability distribution over words in its vocabulary, and a word is picked from the distribution and fed back as an input into the model. This procedure is repeated and halts either when the summary reaches a length constraint, or when the Summarizer produces a special \textit{END} token. See Appendix~\ref{appendix:size_and_initialization} for the model size and initialization used to train the summarization paper.

\subsection{Masking Procedure}
\label{section:masking_procedure}
The Masking Procedure decides on a set of keywords that are important elements in the document that should be recoverable using a summary. The keywords are replaced with blanks, indirectly indicating which information should be present in the summary.
We use a tf-idf-based approach to decide on the set of masked keywords, as it is both simple and has been shown to represent word relevance to a document \cite{Ramos2003UsingTT}. Masking procedure implementation details are presented in Section~\ref{appendix:masking} of the Appendix.

We select the k words with highest tf-idf score for the document to serve as the masked words. The k parameter represents a balance: if too many words are masked, the filling-in becomes impossible, but if too few are masked, the Summarizer model will not be encouraged to include sufficient content in its summary. Varying the value of $k$ (10,12,15,20) yielded only small discernible difference in the Summarizers produced, and we use $k=15$ in all our final experiments. 

The masking procedure can be adapted to a specific domain. For instance, if summarizing financial documents, the masking procedure could systematically mask all numbers, encouraging the Summarizer model to add numbers to its summary.

\subsection{Coverage Model}
\label{section:coverage_model}

\begin{figure}
    \centering
    \includegraphics[width=0.5\textwidth]{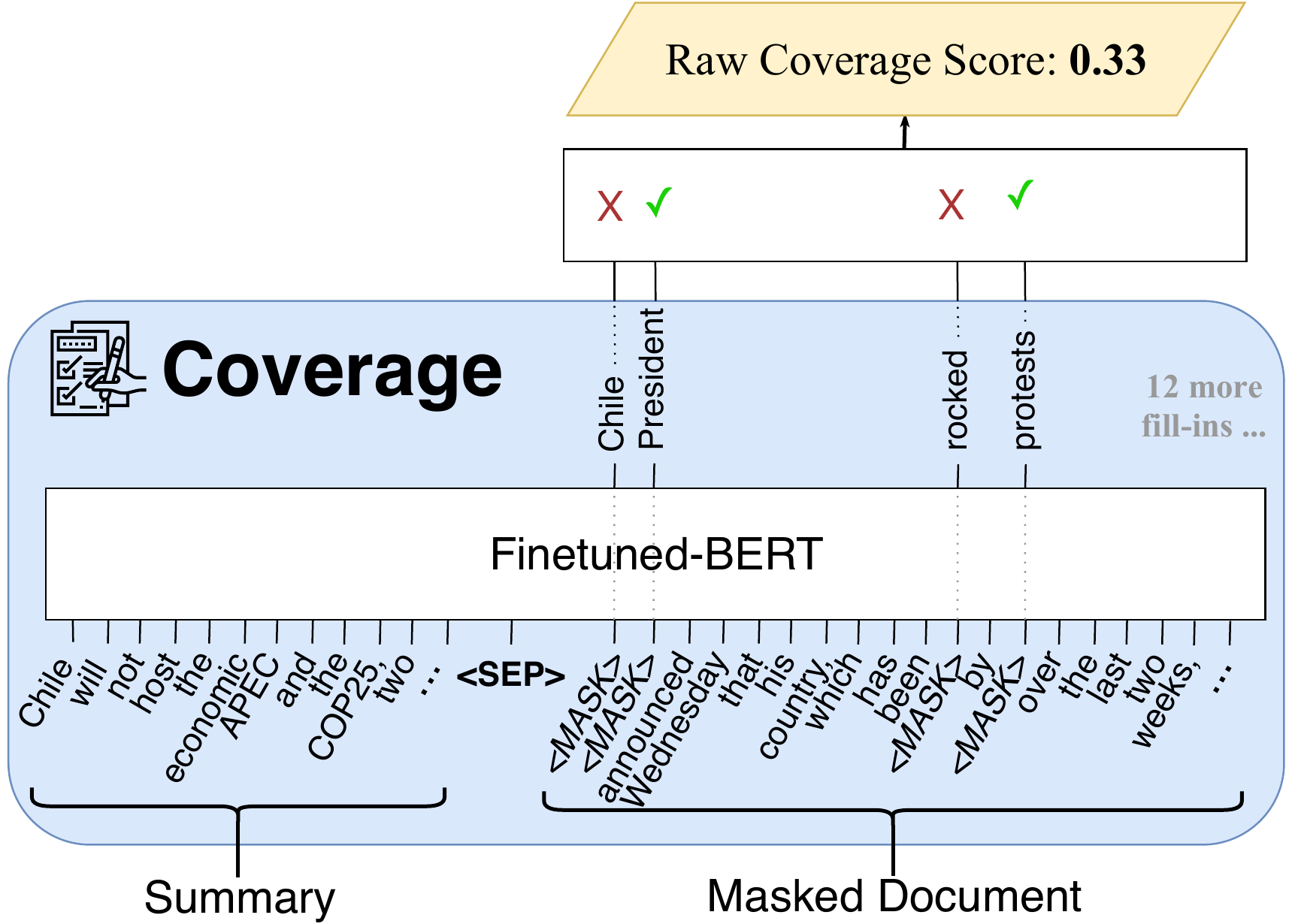}
    \caption{\textbf{The Coverage model uses a finetuned BERT model.} The summary is concatenated to the masked document as the input, and the model predicts the identity of each blank from the original document. The accuracy obtained is the \textit{raw coverage score}.}
    \label{fig:coverage_diagram}
\end{figure}

The Coverage Model receives a computationally generated summary and the masked document and attempts to fill in each blank word. The task of filling in blanks is similar to masked language modeling (MLM), used to pretrain BERT-like \cite{devlin2019bert} models. In MLM, some of the words are replaced with a special $MASK$ token, and the model must use other information (unmasked words) to fill in the masked words.
Because of the similarity to our task, we use a BERT-based neural network as the architecture for the coverage model. However, the coverage task differs from MLM in two ways. First, we modify the masking procedure: instead of masking a random percentage of the words (often 15\% for BERT), we mask all appearances of the keywords selected by the masking procedure described in Section~ \ref{section:masking_procedure}. Second, the input to the coverage model is a concatenation of the unmasked summary, a separator token and the masked document. The model can leverage unmasked information available in the summary to fill in the masked document. The Coverage Model is illustrated in Figure~\ref{fig:coverage_diagram}.

\subsubsection{Computing a Coverage Score}

Using the masking procedure, we obtain $M=f(D)$, the masked document. The coverage model produces the filled document $F=g(M,S)$. \textit{Raw coverage score} is the fraction of correctly filled in words in F. Let $D_i, F_i$ and $M_i$ correspond to the $i$th word in their respective document, $I_M$ the set indices of words that have been masked. Then:
\begin{equation}
    \RawCov(D, S) = \frac{\|i \in I_M \text{ if } D_i = F_i \|}{\| I_M \|}
    \label{eqn:raw_coverage}
\end{equation}

The model can use information in the unmasked (visible) words of M to predict the masked words.  For instance, if the word ``Chile'' is visible, then ``Santiago'' would be a well-informed guess near the word ``capital'', which might not be masked out. This is undesirable, because coverage should account for what information the model can learn from the summary S, not what it can guess from the unmasked portion of D. To counteract this problem, we modify the raw coverage score by computing how much information the model can guess without the summary present, using an empty string summary: $F_\emptyset = g(M, \text{ `` ''})$. We then normalize a summary's coverage by subtracting the empty string coverage from the raw coverage, leaving only filled-in words answerable using S, as shown in Equation \ref{eqn:norm_coverage}.
\begin{equation}
    \begin{aligned}
    &\NormCov(D, S) = \\
    &\RawCov(D, S) - \RawCov(D, \text{`` ''})
    \end{aligned}
    \label{eqn:norm_coverage}
\end{equation}

In a nutshell, raw coverage score answers the question: ``What fraction of blanked words can be correctly filled in with this summary?'' and normalized coverage score answers: ``What is the increase in the fraction of blanks that can be correctly filled in with this summary, compared to having no summary?'' In the rest of this paper, Coverage Score refers to Normalized Coverage Score.

\begin{table}[]
    \resizebox{0.5\textwidth}{!}{%
    \begin{tabular}{rccc}
    \textbf{\begin{tabular}[c]{@{}r@{}}Summary\\ Dataset\end{tabular}} & \textbf{\begin{tabular}[c]{@{}c@{}}Summary\\ Length\end{tabular}} & \textbf{\begin{tabular}[c]{@{}c@{}}Raw\\ Coverage\end{tabular}} & \textbf{\begin{tabular}[c]{@{}c@{}}Norm.\\ Coverage\end{tabular}} \\ \hline
    \textbf{Empty String} & 0 & 0.334 & 0 \\ \hline
    \textbf{Headline} & 9.59 & 0.478 & 0.144 \\
    \textbf{First 10 words} & 10.0 & 0.428 & 0.094 \\ \hline
    \textbf{Newsroom} & 23.41 & 0.525 & 0.191 \\
    \textbf{First 24 words} & 24.0 & 0.537 & 0.203 \\ \hline
    \textbf{CNN/DM} & 45.75 & 0.726 & 0.392 \\ 
    \textbf{First 46 words} & 46.0 & 0.649 & 0.315 \\ \hline
    \end{tabular}
    }
    \caption{Analysis of the raw and normalized coverage of three existing human-written summary datasets, as well as first-k word baselines.}
    \label{fig:coverage_analysis}
\end{table}

\subsubsection{Training the Coverage Model}
\label{section:coverage_pretraining}
We train the Coverage Model once, and its weights are then fixed during the training of the Summarizer.
In order to train the Coverage Model, we need pairs of documents (D) and summaries (S). However, we operate under the assumption that we do not have access to summaries (to keep the procedure unsupervised). In order to remove this dependency, we use the first 50 words  of the unmasked document ($D[:50]$) as a proxy for document summaries.
The Coverage Model is initialized with a trained BERT model \cite{devlin2019bert}, and trained using $(D, D[:50])$ pairs on the coverage task. Because BERT is already trained on the similar MLM task, the Coverage model is able to leverage knowledge accrued by BERT. The Coverage Model converges after roughly 5 hours of training on a Titan X GPU.

\subsubsection{Analysis of Coverage}
We present properties of the raw and normalized coverage through the analysis of existing human-written summary datasets. We focus our analysis on three datasets in the news domain: (1) a headline dataset obtained from common US news websites \cite{laban2017newslens}, (2) the Newsroom dataset  \cite{grusky2018newsroom}, and (3) the CNN/DM dataset \cite{nallapati2016abstractive}.

For each dataset, we take document/summary pairs and obtain raw and normalized coverage score through our Coverage model, reported in Table~\ref{fig:coverage_analysis}.

First, longer summaries obtain higher coverage scores: a CNN/DM summary with an average of 45 words can be used to fill in 73\% of the blanks correctly, compared to 48\% for a 9 word headline. Across datasets, the correlation between summary length and raw coverage score is 0.56, confirming that longer summaries contain more information, according to coverage.

Second, we simulate the first k words\footnote{We choose the first k words due to the similarity to Lede 3 (first 3 sentences), a common baseline in news.} of the document as a summary. We use $k = 10, 24, 46$ to match average word length in the three datasets. For two of the three values (10 and 46), the coverage of human-written summaries is higher than the first-k word counterpart. This is remarkable: even though the summary is farther away lexically (i.e., is not a subset of the original words), it obtains higher coverage, demonstrating that the coverage model can account for reworded information.

\subsection{Fluency Model}
\label{section:fluencymodel}

A model solely trained to optimize coverage has no incentive to write in good English, use punctuation, determinants or pronouns, as these are not words removed by the masking procedure. The objective of a Fluency Model is to judge  the writing quality of the summary, independent of its coverage.

Given the right corpus, we argue that a language model's probability can be modified into a Fluency Score. Therefore, we adapt a language model into the Fluency Model.

We choose the generative Transformer \cite{radford2019language} architecture for our Fluency model, as it can be trained into a powerful language model. Just as with the Summarizer, by using a standardized architecture and model size, we can make use of pretrained models. However, it is important for Fluency to fine tune the language model on the target domain, so that the Summarizer is rewarded for generating text similar to target content.

To produce a uniform Fluency Score, we linearly scale the language model's log-probability of a given summary ($LM(S)$) between an ideal value $LP_{low}$ and a maximum value $LP_{high}$:
\begin{equation}
    \Fluency(S) = 1- \frac{LM(S) - LP_{low}}{LP_{high} - LP_{low}}
    \label{eqn:fluency}
\end{equation}
This ensures that the $\Fluency(S)$ is usually in the range $[0,1]$. $LP_{low}$ and $LP_{high}$ are picked specifically for a particular language model, and ensure that the log-probability magnitudes of a specific language model do not affect the overall scores.

\subsection{Summary Score}

The final Summary Score is a weighed sum of the Coverage and Fluency Scores:
\begin{equation}
    \begin{aligned}
    &\SummaryScore(D,S) =\\
    &\alpha \cdot \NormCov(D,S) + \beta \cdot \Fluency(S)
    \end{aligned}
\end{equation}
$\alpha, \beta$ are hyperparameters giving relative importance to Coverage and Fluency. We set $\alpha = 5$, $\beta=1$ in all our experiments. Model choice, size, and initialization are summarized in Figure~\ref{fig:model_initialization}.

\section{Training Procedure}
\label{section:training}
We first outline the training procedure and then detail several guard-rail mechanisms used during training to prevent the Summarizer from learning pathological writing strategies. 
Figure~\ref{figure:training_plots} presents training plots of a Summary Loop model and interpretation of the different learning phases.

\subsection{Training with Reinforcement Learning}
We use Reinforcement Learning to train the Summarizer component (agent), such that it achieves high summary score (reward). Note that the Coverage and Fluency models are frozen, and their weights are not trained. We make this choice as allowing Fluency and Coverage models to evolve could enable the models to coordinate and cheat.

We use the Self-critical sequence training (SCST) method \cite{rennie2017self}, as it has been shown to perform well on similar text generation tasks optimizing BLEU for image captioning or ROUGE scores in summarization.

In SCST, the Summarizer is used to produce two summaries of document $D$: a greedy summary $\hat{S}$, using a decoding strategy that always picks the most likely next word, and a sampled summary $S^s$, picking the next word in the summary by sampling from the word distribution.

Summaries are scored using the Summary Loop:
\[
    \vspace{-1em}
    \hat{R} = SummaryScore(D,\hat{S})
\]
\[
    R^s = SummaryScore(D,S^s)
\]

Then we minimize the following loss:
$$L = (\hat{R} - R^s) \sum_{i=0}^{N} \log p(w_i^s | w_1^s, ..., w_{i-1}^s, D)$$
Where $p(w_i^s | ...)$ represent the probability of the $i$th word conditioned on previously generated word, according to the model.

Intuitively, if $R^s > \hat{R}$, minimizing L maximizes the likelihood of the sampled sequence \textemdash~which is desired because it outperformed the greedy summary \textemdash~and increases expected reward of the model. 

\subsection{Training guard rails}
\begin{table*}[]
    \centering
    \resizebox{0.99\textwidth}{!}{%
    \begin{tabular}{llllccc}
         \hline
         \textbf{Method} & \textbf{R-1} & \textbf{R-2} & \multicolumn{1}{l|}{\textbf{R-L}} & \multicolumn{1}{l}{\textbf{\begin{tabular}[c]{@{}l@{}}Coverage\\ Score\end{tabular}}} & \multicolumn{1}{l}{\textbf{\begin{tabular}[c]{@{}l@{}}Fluency\\ Score\end{tabular}}} & \multicolumn{1}{l}{\textbf{\begin{tabular}[c]{@{}l@{}}Brevity\\ (avg words)\end{tabular}}} \\ \hline
         \multicolumn{7}{c}{\textbf{Baselines}} \\ \hline
         Human-written Summaries & 100 & 100 & 100 & 0.392 & 0.612 & 58.5 \\
         $\mathds{X}$ Lead-3 baseline & 40.3 & 17.7 & 36.6 & \textbf{0.421} & \textbf{0.656} & \textbf{84.0} \\ \hline
         \multicolumn{7}{c}{\textbf{Supervised Methods}} \\ \hline
         Pointer Generator \cite{see2017get} & 36.4 & 15.7 & 33.4 & 0.342 & 0.547 & 55.6 \\
         PG + Coverage \cite{see2017get} & 39.5 & 17.3 & 36.4 & 0.377 & 0.508 & 61.7 \\
         Bottom-Up \cite{gehrmann2018bottom} & 41.2 & 18.7 & 38.3 & 0.378 & 0.538 & 73.9 \\
         $PEGASUS_{BASE}$ \cite{zhang2019pegasus} & 41.8 & 18.8 & 38.9 & - & - & - \\ 
         $PEGASUS_{LARGE}$ \cite{zhang2019pegasus} & \textbf{44.1} & \textbf{21.3} & \textbf{40.9} & - & - & - \\ \hline
         \multicolumn{7}{c}{\textbf{Unsupervised Methods}} \\ \hline
         $\mathds{X}$ TextRank \cite{mihalcea2004textrank} & 35.2 & 12.9 & 28.7 & 0.370 & 0.612 & 49.62 \\
         GPT2 Zero-Shot \cite{radford2019language} & 29.3 & 8.3 & 26.6 & - & - & - \\
         Summary Loop 45 & \textbf{37.7} & \textbf{14.8} & \textbf{34.7} & 0.404 & 0.627 & 47.0 \\ \hline
         \end{tabular}
    }
    \caption{ROUGE Results (F-1) on the non-anonymized CNN/DM test-set for supervised and unsupervised methods. Extractive methods indicated with $\mathds{X}$. Our ROUGE scores have a 95$\%$ confidence interval of at most $\pm$0.30. Coverage, Fluency and Brevity (average number of words) included for systems where summaries are available, using Coverage and Fluency models from our work.}
    \label{table:cnndm_rouge}
\end{table*}

During training, the Summarizer model learns pathological summarization strategies. We build training guard rails to detect the pathological behavior and penalize the model during training.

A guard rail has a binary effect: if a pathology is detected in a summary, its Summary Score is reduced by a penalty amount $\delta$. We use $\delta=2$ for all experiments. We found three training guard rails to be useful: No-repetition, Finish-your-sentence, and No-frame-filling.

\subsubsection{No-repetition}

A common problem in neural text generation is repetition of text. Based on the observation that 3-grams seldom repeat in common summarization datasets, the ``No-repetition'' training guard rail raises a penalty on a summary when it contains any repeated 3-gram.

\subsubsection{Finish-your-sentence}

When generating a summary, the model can either produce the END token, or generate a number of words up to the length constraint. We observe that if the model does not produce the END token, it often generates partial sentences, which is undesirable. Because we want to encourage the model to generate an END token, the ``Finish-your-sentence'' raises a penalty if a summary has no END token.

\subsubsection{No-frame-filling}

During training, the model sometimes learns to overly rely on sentence patterns that achieves high reward as a one size fits all summary. In one example the model learns to produce summaries solely of the form: ``X talks with Y about the Z''. The model uses this frame, filling in the X, Y and Z slots with relevant keywords and entities to achieve a small but positive coverage. This form of frame-filling is undesirable, as the model often produces inaccurate information to fit the entities to the pattern.

We implement a guard rail to penalize the model when frame-filling patterns are observed. During training, we keep track of the last 100 summaries produced by the model. We then aggregate the frequency of words for each word position in the 100 summaries. If any word appears more than 50\% of the time at a specific word position, we raise the ``No-frame-filling'' penalty. In the example given above, the word ``talks'' appeared in the second word position in more than 50\% of the summaries, as well as the word ``about'' in the fifth position.
% END OF NEW DESCRIPTION

These rule-based training guard rails are simple and effective. In our finalized trained models, very few summaries exhibit penalized behavior: 2\% for no-repetition, 5\% for finish-your-sentence, and 2.5\% for no-frame-filling.

\section{Results}
\label{section:results}
We present results for Summary Loop models trained in the news domain under three different length constraints: 10, 24, and 46 words, matching the distributions of the Headline, Newsroom \cite{grusky2018newsroom} and CNN/DM \cite{nallapati2016abstractive} datasets. We compare our summaries using the standard ROUGE metric, and by analyzing summaries for the errors made, the technique used and the level of abstraction. Finally, we show the Summary Loop can be complemented with supervision, reducing the amount of data needed to achieve comparable ROUGE results.

\subsection{News ROUGE Scores}

\begin{table}[]
    \begin{tabular}{llll}
    \hline
    \textbf{Supervised Methods}   & \textbf{R-1} & \textbf{R-2} & \textbf{R-L} \\ \hline
    $\mathds{X}$ Lead-3 baseline             & \textbf{32.0}        & \textbf{21.1}        & \textbf{29.6}        \\
    PG + Coverage                 & 27.5        & 13.3        & 23.5        \\ \hline
    \textbf{Unsupervised Methods} & \textbf{R-1} & \textbf{R-2} & \textbf{R-L} \\ \hline
    $\mathds{X}$ TextRank                    & 24.5        & \textbf{10.1}        & 20.1        \\
    Summary Loop 24 & \textbf{27.0}        & 9.6         & \textbf{26.4}        \\ \hline
    \end{tabular}
    \caption{ROUGE Results on the released test set of Newsroom. $\mathds{X}$ indicate extractive methods. Summary Loop outperforms other unsupervised method, is competitive with supervised Pointer-Generator.}
    \label{table:newsroom_rouge}
\end{table}

Table~\ref{table:cnndm_rouge} and Table~\ref{table:newsroom_rouge} present ROUGE results on the CNN/DM and Newsroom datasets respectively. In both cases, Summary Loop outperforms other unsupervised methods, and is competitive with supervised methods despite not being exposed to  any example summaries. On CNN/DM, Summary Loop performs in between the Pointer Generator and Bottom Up architecture in terms of ROUGE-1. On the Newsroom, Summary Loop is within 0.6 ROUGE-1 points of the Pointer-Generator with Coverage and surpasses it by 2 ROUGE-L points.

Recent breakthroughs in pretrained Transformer models have shown that using larger models in Summarization can lead to large improvements. For instance, a ``large'' version of the PEGASUS model \cite{zhang2019pegasus} outperforms the ``base'' version by 2.3 ROUGE-1 points. Because Summary Loop experiments were performed using ``base'' models, we expect that using larger Transformer models could lead to similar gains.

Table~\ref{table:cnndm_rouge} confirms that human-written summaries obtain amongst the highest Fluency and Coverage scores. Human-written summaries are only outperformed by Summary Loop summaries, and the Lede-3 baseline. However, the Summary Loop summaries are obtained by directly optimizing for Fluency and Coverage, and Lede-3 baseline summaries achieve their higher Coverage at the expense of being much longer (i.e. 84 words on average compared to 58 in human-written summaries).

\subsection{Technique and Error Analysis}
\label{section:technique_and_error}

\begin{table}[]
    \resizebox{0.5\textwidth}{!}{%
    \begin{tabular}{lccc}
    \hline
    \textbf{Error Made}                                                                & \textbf{PGC} & \textbf{BU} & \textbf{SL} \\ \hline
    Inaccurate (\%)    & \textbf{11} & 31 & 24 \\
    Ungrammatical (\%) & \textbf{7} & 15 & 18 \\ \hline
    \textbf{\begin{tabular}[c]{@{}l@{}}Technique Used\\ (Success/Total)\end{tabular}} & 
    \textbf{\begin{tabular}[c]{@{}c@{}}PGC\\ (S/T)\end{tabular}} &
    \textbf{\begin{tabular}[c]{@{}c@{}}BU\\ (S/T)\end{tabular}} & 
    \textbf{\begin{tabular}[c]{@{}c@{}}SL\\ (S/T)\end{tabular}}\\\hline
    %\textbf{\begin{tabular}[c]{@{}l@{}}Technique Used\\ (success/total)\end{tabular}} & \textbf{PGC} & \textbf{BU} & \textbf{SL} \\ \hline
    Sent. Compression   & 86 / 110 & 96 / 177 & 118 / 194 \\
    Sent. Merging       & 13 / 27 & 29 / 65 & 71 / 121     \\
    Novel Sentence      & 0 / 1  & 4 / 18   & 33 / 70     \\
    Entity Manipulation & 7 / 10 & 15 / 27  & 27 / 40\\
    \textbf{Total Technique} & 106 / 148 & 144 / 287 & 249 / 425 \\
    \hline
    \end{tabular}
    }
    \caption{Error and Technique analysis on 200 randomly selected summaries on the CNN/DM test-set for the Point-Gen with Cov. (PGC), Bottom-Up (BU) and unsupervised Summary Loop (SL). For each summarization technique, we report two numbers: the number of successful occurrences in summaries with no error, and the total number of occurrences in the 200 summaries.}
    \label{table:error_and_technique_analysis}
\end{table}

We perform a manual analysis of 200 randomly-selected summaries on the test set of CNN/DM from the Pointer-Generator with Coverage (PGC), Bottom-Up (BU) and the unsupervised Summary Loop (SL). We annotated each summary with two types of errors: Inaccurate (information in summary contradicts document), Ungrammatical (one sentence or more is not properly constructed), and four summarization techniques: Sentence Compression (summary sentence is a document sentence with words removed), Sentence Merging (2 or more document sentences are merged into a summary sentence), Novel Sentence (original sentence in the summary), and Entity Manipulation (a named entity is modified or simplified, e.g. changing a full name to a last name). We present Summary Loop examples illustrating each error and technique in Figures~\ref{fig:extra_examples1} \textendash~\ref{fig:extra_examples6}.

The analysis was performed by the first author of the paper, labeling article/summary pairs without knowledge of model origin. A summary can manifest any number of summarization Techniques, or none. Labeling is binary: if a summary exhibits more than one or instances of a Technique, it receives a 1, otherwise it receives a 0. Results of the analysis are summarized in Table~\ref{table:error_and_technique_analysis}.

SL uses significantly more summarization techniques (425) than PGC (148) and BU (287) summaries. Beyond raw counts, SL is more successful at applying summarization techniques (59\% success) than BU (50\% success), but less successful than PGC (72\%). Note however that PGC takes little risk: 19\% of the summaries go beyond sentence compression, and 39\% are extractive, using none of the summarization techniques.

\subsection{Level of Abstraction}
\begin{figure}[t]
    \centering
    \pdftooltip{\includegraphics[width=0.47\textwidth]{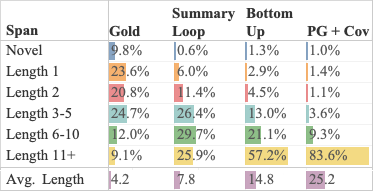}}{Histogram of lengths of copied spans in abstractive summaries. Gold summaries: (novel: 9.8\%, Length 1: 23.6\%, Length 2: 20.8\%, Length 3-5: 24.7\%, Length 6-10: 12.0\%, Length 11+: 9.1 \%. Average Lengte: 4.2). Summary Loop: (0.6\%, 6.0\%, 11.4\%, 26.4\%, 29.7\%, 25.9\%, 7.8). Bottom up: (1.3\%, 2.9\%, 4.5\%, 13.0\%, 21.1\%, 57.2\%, 14.8). PG+Cov: (1.0\%, 1.4\%, 1.1\%, 3.6\%, 9.3\%, 83.6\%, 25.2).}
    
    \caption{Histogram and average copied span lengths for abstractive summaries. A summary is composed of novel words and word spans of various lengths copied from the document. Summary Loop summaries copy shorter spans than prior automatic systems, but do not reach abstraction levels of human-written summaries.}
    \label{table:abstraction_analysis}
\end{figure}

All methods generating summaries one word at a time have potential for abstraction. In Figure~\ref{table:abstraction_analysis} we analyze human and system written summaries for abstraction level. We measure  a summary's level of abstraction by looking at the length of spans copied from the document. Summary Loop is the most abstractive automated method, although less so than human written summaries.  SL cuts nearly in half the length of copied spans compared to other automated methods.

\subsection{Supervision is not the enemy}
\begin{table}[]
    \resizebox{0.48\textwidth}{!}{%
	\begin{tabular}{lllll}
	\hline
	Initialization Method & R-1 & R-2 & R-L & Test Loss \\ \hline
	\multicolumn{5}{c}{28k samples from CNN/DM (10\%)} \\ \hline
	Random Initialization & 7.0 & 0.9 & 8.8 & 6.05 \\
	GPT2 & 37.1 & 15.9 & 31.9 & 2.21 \\
	Summary Loop S10 & \textbf{38.7} & \textbf{16.2} & \textbf{35.1} & 2.07 \\ \hline
	\multicolumn{5}{c}{All of CNN/DN (100\%)} \\ \hline
	Random Weights & 20.4 & 4.1 & 19.1 & 4.22 \\
	GPT2 & 38.4 & 17.2 & 35.0 & 2.02 \\
	Summary Loop S100 & \textbf{41.0} & \textbf{18.1} & \textbf{37.3} & 1.89 \\\hline
	\end{tabular}
	}
	\caption{ROUGE Results on the CNN/DM test-set for supervised generative Transformers. Initializing with the unsupervised Summary Loop outperforms random and GPT2 initializations.}
	\label{table:supervision_results}
\end{table}

If summaries are available, we show that they can complement the unsupervised Summary Loop. We run supervised experiments on CNN/DM using a generative Transformer architecture and varying the initialization. We compare initializing with (1) random weights, (2) the original GPT2 weights, and (3) the Summary Loop weights of target length 45. We train each model with teacher forcing, comparing using the entire CNN/DM training set to just 10\% of it. The results are summarized in Table~\ref{table:supervision_results}.

First, initializing with the Summary Loop leads to higher ROUGE score both in the 10\% and full dataset setting. As expected, results improve when using the entirety of the data, and the Summary Loop initialized model trained with the entirety of CNN/DM obtains a ROUGE-1 F1-score of 41.0, within the confidence interval of the supervised Bottom Up \cite{gehrmann2018bottom} architecture. This is a strong result as the Transformer we use is a generic language model, and is not specialized for summarization.

Second, initializing with Summary Loop and training with 10\% of CNN/DM yields comparable ROUGE scores to initializing with GPT2 and using the entire CNN/DM, showing that Summary Loop can be useful when fewer summaries are available.

\section{Discussion}

\textbf{Customizing summaries.} In Figure~\ref{fig:coverage_example}, we illustrate the effect of the length constraint by summarizing the same document under three different length constraints. Each model adapts to its word budget. However, length is only one way to customize summaries. One might want to summarize based on point of view, chronology, theme, etc.

\textbf{Fluency vs. Grammaticality.} By choosing to represent the validity of summaries with a Language model, we encourage fluent summaries (i.e., with likely sequences of words) but not necessarily grammatical ones. Extending the scoring to include grammaticality, either by using a parsing model, or leveraging the Corpus of Linguistic Acceptability \cite{warstadt2019neural} could prove useful.

\textbf{Summarization in the wild.} Because our method is unsupervised, it can be applied to new domains and languages. In this work, we benefited from pretrained BERT and GPT2 models in English, which do not yet exist publicly for other languages. Once they become available in other languages, the Summary Loop can be ported over.

\textbf{Abstraction dangers.}  Recent work around measuring factuality in generated text, using Natural Language Inference \cite{guo2018soft} or rule-based fact extraction \cite{zhang2019optimizing} becomes increasingly important with summaries that are more abstractive. This work can be naturally included into the Summary Loop, with a fact-checker model generating an accuracy score.

\section{Conclusion}

In this work we present a new approach to unsupervised abstractive summarization based on maximizing a combination of coverage and fluency for a given length constraint. When tested on common news summarization datasets, our method significantly outperforms previous unsupervised methods, and gets within the range of competitive supervised methods. Our models attain levels of abstraction closer to human-written summaries, although with more abstraction, more potential for factual inaccuracies arise.

\section*{Acknowledgments}

We would like to thank Forrest Huang, David Chan, Roshan Rao, Katie Stasaski and the ACL reviewers for their helpful comments. This work was supported by the first author's internship at Bloomberg, and a Bloomberg Data Science grant. We also gratefully acknowledge support received from an Amazon Web Services Machine Learning Research Award and an NVIDIA Corporation GPU grant.

\bibliography{anthology,acl2020}
\bibliographystyle{acl_natbib}
\clearpage
\appendix
\renewcommand{\thetable}{A\arabic{table}}
\setcounter{table}{0}
\renewcommand{\thefigure}{A\arabic{figure}}
\setcounter{figure}{0}

\section{Masking Procedure Details}
\label{appendix:masking}
The masking procedure follows these steps:
\begin{enumerate}
    \item We randomly sample 5,000 documents in the domain being summarized (e.g. News) as a training corpus,
    \item The training corpus is tokenized using the tokenizer of the Coverage model. In our case, we tokenize with the Word Piece model of the BERT Base model \cite{devlin2019bert},
    \item We train a tf-idf transformation model using the tokenized training corpus using default parameters of scikit-learn's tf-idf implementation \cite{scikit-learn},
    \item Given a document to be masked, we use the trained tf-idf model to produce a tf-idf for the document,
    \item The words present in the document are ranked in decreasing order of tf-idf score, and the $k$ words with highest tf-idf form the masking set,
    \item All occurrences of the words in the masking set are replaced by a mask in the document, creating the masked document.
\end{enumerate}

\section{Fluency Examples}
Table~\ref{fig:fluency_examples} provides examples from the Headline dataset of sampled headlines and their corresponding Fluency Score. The Fluency Score, a normalized language model log-perplexity, ranges from 0 to 1. Even though all these headlines are written by a human, the Fluency scores vary, with the higher-scoring headlines using more standard grammatical constructs. Note that the use of complex entity names does not prevent the model from obtaining a high Fluency score.

\begin{table}[!htbp]
    \resizebox{0.5\textwidth}{!}{%
    \begin{tabular}{p{5.8cm} c}
    \hline
    Example Headline                                                       & Fluency Score \\ \hline
    Henry's Monaco recruit giant Brazilian Naldo for relegation scrap      & 0.16          \\
    Tesla shares dive after price cut, production numbers                  & 0.41          \\
    French police arrest gilets jaunes protests leader Eric Drouet         & 0.59          \\
    Carlos Ghosn will appear in public for the first time since his arrest & 0.75          \\ \hline
    \end{tabular}
    }
    \caption{\textbf{Example selected headlines and their Fluency score.} The headlines were picked from a corpus of human-written news headlines. The average Fluency in the corpus is 0.479.}
    \label{fig:fluency_examples}
\end{table}

\section{Model Size and Initialization}
\label{appendix:size_and_initialization}
 Figure~\ref{fig:model_initialization} shows the model size and initialization model used  for each of the Summarizer, Coverage and Fluency models.
\begin{figure}[!htbp]
    \begin{framed}
        \vspace{-0.5em}
        \begin{center}
            \textbf{Summarizer Architecture}
        \end{center}
        GPT2-base: 12-layer, 768-hidden, 12-heads
        \begin{center}
            \textbf{Summarizer Initialization}
        \end{center}
        GPT2 base model from \citet{radford2019language}
        
        \vspace{0.5em}
        \hrule
        
        \begin{center}
            \textbf{Coverage Architecture}
        \end{center}
        BERT-base: 12-layer, 768-hidden, 12-heads
        \begin{center}
            \textbf{Coverage Initialization}
        \end{center}
        Pretrained model obtained in Section~\ref{section:coverage_pretraining}
    
        \vspace{0.5em}
        \hrule
    
        \begin{center}
            \textbf{Fluency Architecture}
        \end{center}
        GPT2-base: 12-layer, 768-hidden, 12-heads
        \begin{center}
            \textbf{Fluency Initialization}
        \end{center}
        GPT2 base model from \cite{radford2019language}, finetuned with Language modeling on news text.
        \vspace{-1.0em}
    \end{framed}
    \caption{The model size choice as well as initialization method for the Summarizer, Coverage and Fluency models in the Summary Loop. Each model leverages a pretrained Transformer.}
    \label{fig:model_initialization}
\end{figure}

\section{Training Plots}

\begin{figure*}%
    \centering
    \subfigure[Fluency Score]{%
        \label{fig:fluency_plot}%
        \includegraphics[width=0.47\textwidth]{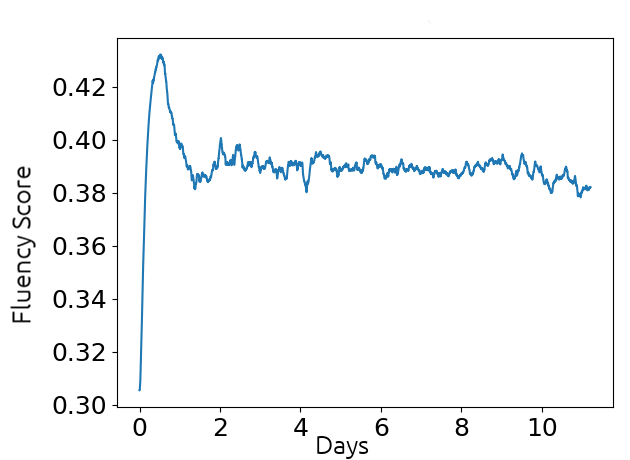}}%
    \qquad
    \subfigure[Coverage Score]{%
        \label{fig:coverage_plot}%
        \includegraphics[width=0.47\textwidth]{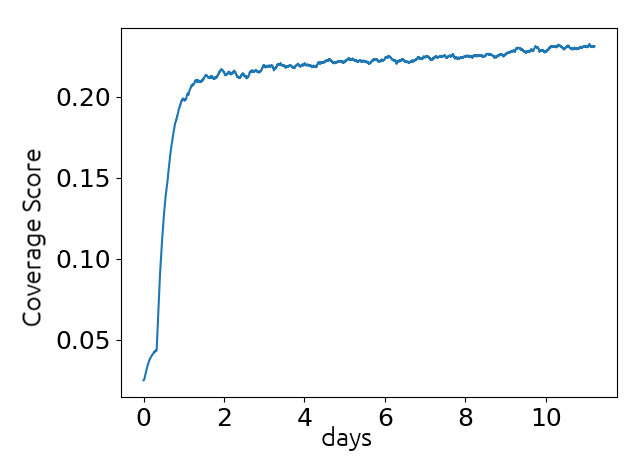}}%
    \qquad
    \subfigure[Summary Score]{%
        \label{fig:total_plot}%
        \includegraphics[width=0.47\textwidth]{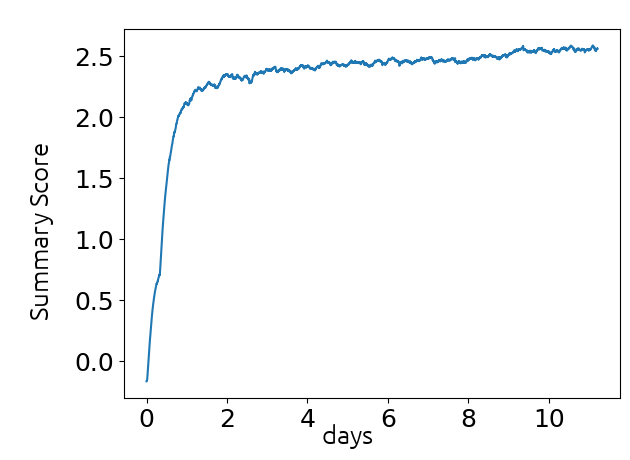}}%
    \qquad
    \subfigure[Average number of words in summary]{%
        \label{fig:nword_plot}%
        \includegraphics[width=0.47\textwidth]{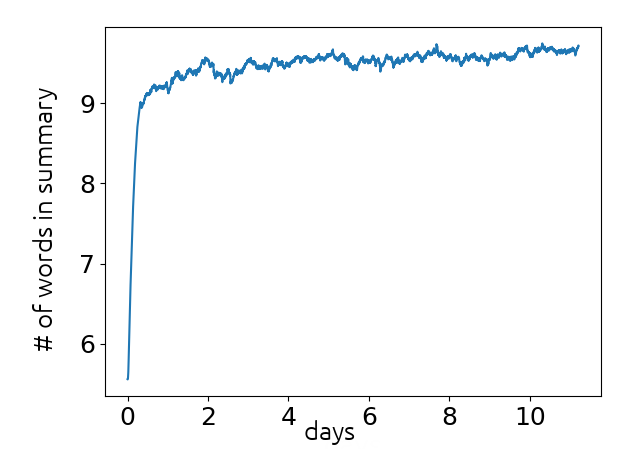}}%

    \caption{Plots of key variables during the training of the length 10 Summary Loop: \subref{fig:fluency_plot} is a plot of the average Fluency Score, \subref{fig:coverage_plot} is a plot of the average normalized Coverage Score, \subref{fig:total_plot} is a plot of the average Summary Score (taking guard-rails into account), and \subref{fig:nword_plot} is a plot of the average number of words in summaries produced.}
    \label{figure:training_plots}
\end{figure*}

Figure~\ref{figure:training_plots} presents the plots of key variables we obtain during the training of the length 10 Summary Loop model. The training occurred over 10 days using a single Titan X GPU. During a first phase which occurs in the first 2 days of training, the model learns to copy content from the news article, which helps it achieve high Fluency and Coverage. In a second phase starting around the second day, the Summarizer learns to gain Coverage which maintaining Fluency mostly constant, which makes the overall Summary Score rise. The Summarizer model quickly learns to use its word budget, and  after 10 days of training, the model uses an average of 9.7 words in its summaries.

\section{Example Annotated Summaries}

Figures \ref{fig:extra_examples1}, \ref{fig:extra_examples2}, \ref{fig:extra_examples3}, \ref{fig:extra_examples4}, \ref{fig:extra_examples5}, and \ref{fig:extra_examples6} show example documents and the generated Summary Loop summary from the error and technique analysis of Section~\ref{section:technique_and_error}. Each summary manifests a summarization technique or error observed.

\begin{figure*}[!htbp]
    \begin{framed}
        % \small
        \begin{center}
            \large
            \textbf{Sentence Compression Example}
        \end{center}
        \textbf{Document:}
        He has long struggled to convince voters that he is a suitable choice for prime minister. Now Ed Miliband has hired a leadership coaching firm that helps people overcome anxiety and find their ``inner voice''. \textbf{\textcolor{blue}{The consultants drafted in by the Labour leader claim to work with politicians}} to build "leadership skills" using ``neuroscience'' and ``business psychology''. Ed Miliband, pictured, has hired a US guru who can help him convince himself that he can be Prime Minister. [...]
        \begin{center}
            ~
        \end{center}
        \textbf{Summary:}
        Ed Miliband has hired a US guru who can help politicians on their leadership skills using neuroscience. Mr Miliband has hired the firm that can help politicians to build their leadership skills. \textbf{\textcolor{blue}{The consultants drafted in by the Labour leader claim to work with politicians.}}
        
    \end{framed}

    \caption{Summary Loop summary from the Error and Technique analysis (Section~\ref{section:technique_and_error}) illustrating the \textbf{Sentence Compression} technique. The blue boldface highlight is an example of sentence compression.}    \label{fig:extra_examples1}
\end{figure*}
\begin{figure*}[!htbp]
    \begin{framed}
        \begin{center}
            \large
            \textbf{Sentence Merging Example}
        \end{center}
        
        \textbf{Document:} 
        A single mom and her three kids who \textbf{\textcolor{blue}{``lost everything but their lives'' in the East Village apartment explosion last week}} are getting an incredible outpouring of support from their fellow New Yorkers. [...] Dr \textbf{\textcolor{blue}{McLean}}, a 58-year-old child psychiatrist in the South Bronx, \textbf{\textcolor{blue}{says she and daughter Rose, 8, and twins James and Annabelle}}, 5, had nothing more than the clothes on their backs after the disaster. \textit{\textcolor{red}{Diane McLean}}, 58, \textit{\textcolor{red}{and her three children}} lost ``everything but their lives'' when fire destroyed their apartment last week. Rose, 8, ( left ) and twins James and Annabelle, 5, lost everything except \textit{\textcolor{red}{the clothes on their backs}} in the fire that destroyed their apartment building. [..] A GoFundMe campaign has raised nearly \$ 90,000. [...]
         \begin{center}
            ~
        \end{center}
        \textbf{Summary:} \textbf{\textcolor{blue}{Diane McLean says she and daughter Rose, 8, and twins James and Annabelle, lost everything but their lives at East Village apartment explosion last week.}} \textit{\textcolor{red}{Diane McLean and her three kids had the clothes on their backs.}} A GoFundMe campaign has raised nearly \$ 90,000. 
    \end{framed}

    \caption{Summary Loop summary from the Error and Technique analysis (Section~\ref{section:technique_and_error}) illustrating the \textbf{Sentence Merging} technique. The bold blue and italicized red selections are two examples of sentence merging. In the blue example ``Dr McLean'' is replaced by ``Diane McLean'' in the summary, an example of entity manipulation.}    \label{fig:extra_examples2}
\end{figure*}
\begin{figure*}[!htbp]
    \begin{framed}
        \begin{center}
            \large
            \textbf{Novel Sentence Example}
        \end{center}

        \textbf{Document:}
        For most of us, the dream of a holiday \textbf{\textcolor{blue}{home}} is one that will probably never be realised. But for the lucky minority with a few extra million in the bank, its seems the world is quite literally your oyster when looking for property around the world. From a Lake Garda mansion with a pool overlooking the water to an Italian villa that looks like a castle and an Antigua retreat with Giorgio Armani as a neighbour, these are some of the most spectacular holiday homes on the market at the moment. \textbf{\textcolor{blue}{On the Lombardy side of Lake Garda}}, this Lionard property is a luxurious villa with one serious waterfront view. Lake Garda. On the Lombardy side of Lake Garda, in northern Italy, lies a \textbf{\textcolor{blue}{luxury villa with a view}} - just several miles north of Brescia. And for \euro{} 18 million ( about \pounds 13 million or $\$$20 million ) it can all be yours. Not only is there a large swimming pool looking out on the water, but also a large deck with plenty of space for sun beds, gazebos and al fresco dining spots, overlooking a 4000 square metre garden. Inside, the house is just as breathtaking. For about 18 million Euros ( or $\$$ 13 million ), the modern home, complete with pool, gazebo, and al fresco dining options, can be yours. [...]
         \begin{center}
            ~
        \end{center}
        \textbf{Summary:}
        \textbf{\textcolor{blue}{The Lake Garda home is a luxury villa with a view on the Lombardy side of Lake Garda.}} This villa with gazebo and al fresco dining options. Inside, the house is just as breathtaking. For about 18 million Euros.
        
    \end{framed}

    \caption{Summary Loop summary from the Error and Technique analysis (Section~\ref{section:technique_and_error}) illustrating the \textbf{Novel Sentence} technique. The first sentence of the summary uses pieces from the original document (in boldface blue) to form a sentence with an alternative but correct meaning.}
    \label{fig:extra_examples3}
\end{figure*}
\begin{figure*}[!htbp]
    \begin{framed}
        \begin{center}
            \large
            \textbf{Entity Manipulation Example}
        \end{center}

        \textbf{Document:} Sipping a glass of glorious red wine which has been carefully aged in a hand-crafted oak barrel is my idea of heaven. [...] A $\$$ 5 bottle has suddenly become $\$$ 12 because the wine has lingered in an oak barrel before bottling. So when I read this week about a new gadget that \textbf{\textcolor{blue}{claims to be able to ``oak age'' wine in hours rather than years}}, my curiosity was seriously roused. The Oak Bottle promises to impart an authentic aged flavour -- a process that can take up to two years -- in just a day or two. Who wouldn't drink to that ? Scroll down for video. TV wine expert Oz Clarke puts to the test this oak bottle that claims to ``oak age'' wine in hours rather than years. The product, which retails at $\$$ 50, is the brainchild of 30-year-old entrepreneur \textbf{\textcolor{blue}{Joel Paglione}}. [...]
         \begin{center}
            ~
        \end{center}
        \textbf{Summary:} \textbf{\textcolor{blue}{Joel Paglione}} said the Oak Bottle promises to be able to oak age wine in hours rather than years. The Oak Bottle promises an authentic aged flavour that can take up to two years. A bottle has been made in an oak barrel.
        
    \end{framed}

    \caption{Summary Loop summary from the Error and Technique analysis (Section~\ref{section:technique_and_error}) illustrating the \textbf{Entity Manipulation} technique. The entity Joel Paglione (in boldface blue) is correctly inserted to represent the company.}
    \label{fig:extra_examples4}
\end{figure*}
\begin{figure*}[!htbp]
    \begin{framed}
        \begin{center}
            \large
            \textbf{Inaccurate Example}
        \end{center}

        \textbf{Document:}
        The traditional cookie cutter wedding no longer exists - new reports suggest Brits are ditching tradition in favour of alternative practices when it comes to getting hitched. Two of the biggest changes are the fact \textbf{\textcolor{blue}{that religious services have fallen out of favour}} and that brides are opting for bold colour schemes for their big day. A new study, which has tracked the decisions of brides and grooms over the past five years interviewed 1,893 newlyweds and compared them to answers they have collated since 2010. Scroll down for video. [...]
         \begin{center}
            ~
        \end{center}
        \textbf{Summary:}
        \textbf{\textcolor{blue}{The new study showed that British couples are opting for religious ceremonies}} when it comes to their big day with services falling from 40 per cent of the past five years. The study showed that couples are opting to holiday in the UK.
    \end{framed}

    \caption{Summary Loop summary from the Error and Technique analysis (Section~\ref{section:technique_and_error}) illustrating the \textbf{Inaccurate} error. The summary inaccurately claims religious ceremonies are increasing, when the document says they are in decline.  Key phrases are highlighted in boldface blue.}
    \label{fig:extra_examples5}
\end{figure*}
\begin{figure*}[!htbp]
    \begin{framed}
        \begin{center}
            \large
            \textbf{Ungrammatical Example}
        \end{center}

        \textbf{Document:} Despite his daughter remaining in a medically induced coma since she was found unresponsive in a bathtub at her Atlanta home in January, singer Bobby Brown told an audience on Saturday night that she is ``awake.''. Bobby was performing at the Verizon Theatre in Dallas when he told the stunned audience that ``Bobbi is awake. She's watching me.'' The singer didn't elaborate on if his daughter had regained consciousness or if he was talking instead about her spirit. After the 46-year-old's comment, his sister Tina posted on Facebook," [...] Whitney Houston's family insists the 22-year-old is not awake and is the same condition she was when she entered the facility. "\textbf{\textcolor{blue}{She's in the exact same condition she was in when she went into the facility.}}" a source told the site [...]
         \begin{center}
            ~
        \end{center}
        \textbf{Summary:} Bobby Brown was performing at the Verizon Theatre in Dallas when Bobbi was awake. He said that Tina posted on Facebook that her daughter was awake. She was the singer. She was going to be awake. \textbf{\textcolor{blue}{She is the same condition.}}
    \end{framed}

    \caption{Summary Loop summary from the Error and Technique analysis (Section~\ref{section:technique_and_error}) illustrating the \textbf{Ungrammatical} error. The last short summary sentence (in boldface blue) is not properly constructed, based on an unsuccessful attempt to compress a sentence in the document (also in boldface blue).}
    \label{fig:extra_examples6}
\end{figure*}

\end{document}

%% file: new_intro.tex
Summarization, or the task of condensing a document's main points into a shorter  document, is important for many text domains, such as headlines for news and abstracts for research papers.  

This paper presents a novel unsupervised abstractive summarization method that generates summaries directly from source documents, without the aid of example summaries.
This  approach simultaneously optimizes for the following important properties of a good summary: 
\begin{itemize}
    \item  \textbf{coverage} of the keywords of the document, 
    \item  \textbf{fluency} of generated language, and 
    \item  \textbf{brevity} of generated summaries.
\end{itemize}

One of the main contributions of this work is a novel method of inducing good coverage of important concepts from the original article. The coverage model we propose takes as input the original document with keywords masked out (see Figure~\ref{fig:coverage_example}). It uses the current best automatically generated summary to try to uncover the missing keywords. The more informative the current summary is, the more successful the coverage model is at guessing the blanked out keywords from the original document. A resulting coverage score is fed back into the training process of the summarization model with the objective of producing  summaries with high coverage.

A second contribution is our unsupervised training procedure for summarization, the \textit{Summary Loop}, which leverages the coverage model as well as a simple fluency model to generate and score summaries. During training, the procedure is conditioned on a desired summary length, forcing the Summarizer model to adapt to a length budget. Figure~\ref{fig:coverage_example} shows Summary Loop summaries obtained for the same document under three different length budgets.

A third contribution is a set of specialized techniques employed during training to guide the model away from pathological behavior. These \textit{guard rails} include a method for reducing repetition, for encouraging the model to complete sentences, and to avoid frame filling patterns.

The models trained through the Summary Loop outperform all prior unsupervised summarization methods by at least 2 ROUGE-1 points on common news summarization datasets (CNN/DM and Newsroom), and achieve within a few points of state-of-the-art supervised algorithms, without ever being exposed to any summaries.
In addition, summaries generated by our method use 50$\%$ more summarization techniques (compression, merging, etc.) than prior automatic work and achieve higher levels of abstraction, reducing by almost half the gap between human-generated summaries and automatic summaries in terms of length of copied spans.

%% file: acl2020.bbl
\begin{thebibliography}{31}
\expandafter\ifx\csname natexlab\endcsname\relax\def\natexlab#1{#1}\fi

\bibitem[{Arumae and Liu(2018)}]{arumae2018reinforced}
Kristjan Arumae and Fei Liu. 2018.
\newblock Reinforced extractive summarization with question-focused rewards.
\newblock In \emph{Proceedings of ACL 2018, Student Research Workshop}, pages
  105--111.

\bibitem[{Barrios et~al.(2015)Barrios, L{\'o}pez, Argerich, and
  Wachenchauzer}]{barrios2015variations}
Federico Barrios, Federico L{\'o}pez, Luis Argerich, and Rosita Wachenchauzer.
  2015.
\newblock Variations of the similarity function of textrank for automated
  summarization.
\newblock In \emph{Argentine Symposium on Artificial Intelligence (ASAI
  2015)-JAIIO 44 (Rosario, 2015)}.

\bibitem[{Chen and Bansal(2018)}]{chen2018fast}
Yen-Chun Chen and Mohit Bansal. 2018.
\newblock Fast abstractive summarization with reinforce-selected sentence
  rewriting.
\newblock In \emph{Proceedings of the 56th Annual Meeting of the Association
  for Computational Linguistics (Volume 1: Long Papers)}, pages 675--686.

\bibitem[{Chi et~al.(2019)Chi, Dong, Wei, Wang, Mao, and Huang}]{chi2019cross}
Zewen Chi, Li~Dong, Furu Wei, Wenhui Wang, Xian-Ling Mao, and Heyan Huang.
  2019.
\newblock Cross-lingual natural language generation via pre-training.
\newblock \emph{arXiv preprint arXiv:1909.10481}.

\bibitem[{Devlin et~al.(2019)Devlin, Chang, Lee, and
  Toutanova}]{devlin2019bert}
Jacob Devlin, Ming-Wei Chang, Kenton Lee, and Kristina Toutanova. 2019.
\newblock Bert: Pre-training of deep bidirectional transformers for language
  understanding.
\newblock In \emph{Proceedings of the 2019 Conference of the North American
  Chapter of the Association for Computational Linguistics: Human Language
  Technologies, Volume 1 (Long and Short Papers)}, pages 4171--4186.

\bibitem[{Edunov et~al.(2019)Edunov, Baevski, and Auli}]{edunov2019pre}
Sergey Edunov, Alexei Baevski, and Michael Auli. 2019.
\newblock Pre-trained language model representations for language generation.
\newblock In \emph{Proceedings of the 2019 Conference of the North American
  Chapter of the Association for Computational Linguistics: Human Language
  Technologies, Volume 1 (Long and Short Papers)}, pages 4052--4059.

\bibitem[{Eyal et~al.(2019)Eyal, Baumel, and Elhadad}]{eyal2019question}
Matan Eyal, Tal Baumel, and Michael Elhadad. 2019.
\newblock Question answering as an automatic evaluation metric for news article
  summarization.
\newblock In \emph{Proceedings of the 2019 Conference of the North American
  Chapter of the Association for Computational Linguistics: Human Language
  Technologies, Volume 1 (Long and Short Papers)}, pages 3938--3948.

\bibitem[{Gehrmann et~al.(2018)Gehrmann, Deng, and Rush}]{gehrmann2018bottom}
Sebastian Gehrmann, Yuntian Deng, and Alexander Rush. 2018.
\newblock Bottom-up abstractive summarization.
\newblock In \emph{Proceedings of the 2018 Conference on Empirical Methods in
  Natural Language Processing}, pages 4098--4109.

\bibitem[{Grusky et~al.(2018)Grusky, Naaman, and Artzi}]{grusky2018newsroom}
Max Grusky, Mor Naaman, and Yoav Artzi. 2018.
\newblock Newsroom: A dataset of 1.3 million summaries with diverse extractive
  strategies.
\newblock In \emph{Proceedings of the 2018 Conference of the North American
  Chapter of the Association for Computational Linguistics: Human Language
  Technologies, Volume 1 (Long Papers)}, pages 708--719.

\bibitem[{Gui et~al.(2019)Gui, Tian, Wang, and Yang}]{gui2019attention}
Min Gui, Junfeng Tian, Rui Wang, and Zhenglu Yang. 2019.
\newblock Attention optimization for abstractive document summarization.
\newblock In \emph{Proceedings of the 2019 Conference on Empirical Methods in
  Natural Language Processing and the 9th International Joint Conference on
  Natural Language Processing (EMNLP-IJCNLP)}, pages 1222--1228.

\bibitem[{Guo et~al.(2018)Guo, Pasunuru, and Bansal}]{guo2018soft}
Han Guo, Ramakanth Pasunuru, and Mohit Bansal. 2018.
\newblock Soft layer-specific multi-task summarization with entailment and
  question generation.
\newblock In \emph{Proceedings of the 56th Annual Meeting of the Association
  for Computational Linguistics (Volume 1: Long Papers)}, pages 687--697.

\bibitem[{Laban and Hearst(2017)}]{laban2017newslens}
Philippe Laban and Marti~A Hearst. 2017.
\newblock newslens: building and visualizing long-ranging news stories.
\newblock In \emph{Proceedings of the Events and Stories in the News Workshop},
  pages 1--9.

\bibitem[{Lewis et~al.(2019)Lewis, Liu, Goyal, Ghazvininejad, Mohamed, Levy,
  Stoyanov, and Zettlemoyer}]{lewis2019bart}
Mike Lewis, Yinhan Liu, Naman Goyal, Marjan Ghazvininejad, Abdelrahman Mohamed,
  Omer Levy, Ves Stoyanov, and Luke Zettlemoyer. 2019.
\newblock Bart: Denoising sequence-to-sequence pre-training for natural
  language generation, translation, and comprehension.
\newblock \emph{arXiv preprint arXiv:1910.13461}.

\bibitem[{Mihalcea and Tarau(2004)}]{mihalcea2004textrank}
Rada Mihalcea and Paul Tarau. 2004.
\newblock Textrank: Bringing order into text.
\newblock In \emph{Proceedings of the 2004 conference on empirical methods in
  natural language processing}, pages 404--411.

\bibitem[{Nallapati et~al.(2016)Nallapati, Zhou, dos Santos, glar
  Gul{\c{c}}ehre, and Xiang}]{nallapati2016abstractive}
Ramesh Nallapati, Bowen Zhou, Cicero dos Santos, {\c{C}}a~glar Gul{\c{c}}ehre,
  and Bing Xiang. 2016.
\newblock Abstractive text summarization using sequence-to-sequence rnns and
  beyond.
\newblock \emph{CoNLL 2016}, page 280.

\bibitem[{Nikolov and Hahnloser(2019)}]{nikolov2019abstractive}
Nikola~I Nikolov and Richard~HR Hahnloser. 2019.
\newblock Abstractive document summarization without parallel data.
\newblock \emph{arXiv preprint arXiv:1907.12951}.

\bibitem[{Pasunuru and Bansal(2018)}]{pasunuru2018multi}
Ramakanth Pasunuru and Mohit Bansal. 2018.
\newblock Multi-reward reinforced summarization with saliency and entailment.
\newblock In \emph{Proceedings of the 2018 Conference of the North American
  Chapter of the Association for Computational Linguistics: Human Language
  Technologies, Volume 2 (Short Papers)}, pages 646--653.

\bibitem[{Paulus et~al.(2018)Paulus, Xiong, and Socher}]{Paulus2018ADR}
Romain Paulus, Caiming Xiong, and Richard Socher. 2018.
\newblock A deep reinforced model for abstractive summarization.
\newblock In \emph{Proceedings of ICLR}.

\bibitem[{Pedregosa et~al.(2011)Pedregosa, Varoquaux, Gramfort, Michel,
  Thirion, Grisel, Blondel, Prettenhofer, Weiss, Dubourg, Vanderplas, Passos,
  Cournapeau, Brucher, Perrot, and Duchesnay}]{scikit-learn}
F.~Pedregosa, G.~Varoquaux, A.~Gramfort, V.~Michel, B.~Thirion, O.~Grisel,
  M.~Blondel, P.~Prettenhofer, R.~Weiss, V.~Dubourg, J.~Vanderplas, A.~Passos,
  D.~Cournapeau, M.~Brucher, M.~Perrot, and E.~Duchesnay. 2011.
\newblock Scikit-learn: Machine learning in {P}ython.
\newblock \emph{Journal of Machine Learning Research}, 12:2825--2830.

\bibitem[{Radford et~al.(2019)Radford, Wu, Child, Luan, Amodei, and
  Sutskever}]{radford2019language}
Alec Radford, Jeffrey Wu, Rewon Child, David Luan, Dario Amodei, and Ilya
  Sutskever. 2019.
\newblock Language models are unsupervised multitask learners.

\bibitem[{Rajpurkar et~al.(2018)Rajpurkar, Jia, and Liang}]{rajpurkar2018know}
Pranav Rajpurkar, Robin Jia, and Percy Liang. 2018.
\newblock Know what you don’t know: Unanswerable questions for squad.
\newblock In \emph{Proceedings of the 56th Annual Meeting of the Association
  for Computational Linguistics (Volume 2: Short Papers)}, pages 784--789.

\bibitem[{Ramos(2003)}]{Ramos2003UsingTT}
Juan~Enrique Ramos. 2003.
\newblock Using tf-idf to determine word relevance in document queries.

\bibitem[{Rennie et~al.(2017)Rennie, Marcheret, Mroueh, Ross, and
  Goel}]{rennie2017self}
Steven~J Rennie, Etienne Marcheret, Youssef Mroueh, Jerret Ross, and Vaibhava
  Goel. 2017.
\newblock Self-critical sequence training for image captioning.
\newblock In \emph{Proceedings of the IEEE Conference on Computer Vision and
  Pattern Recognition}, pages 7008--7024.

\bibitem[{Scialom et~al.(2019)Scialom, Lamprier, Piwowarski, and
  Staiano}]{scialom2019answers}
Thomas Scialom, Sylvain Lamprier, Benjamin Piwowarski, and Jacopo Staiano.
  2019.
\newblock Answers unite! unsupervised metrics for reinforced summarization
  models.
\newblock In \emph{Proceedings of the 2019 Conference on Empirical Methods in
  Natural Language Processing and the 9th International Joint Conference on
  Natural Language Processing (EMNLP-IJCNLP)}, pages 3237--3247.

\bibitem[{See et~al.(2017)See, Liu, and Manning}]{see2017get}
Abigail See, Peter~J Liu, and Christopher~D Manning. 2017.
\newblock Get to the point: Summarization with pointer-generator networks.
\newblock In \emph{Proceedings of the 55th Annual Meeting of the Association
  for Computational Linguistics (Volume 1: Long Papers)}, volume~1, pages
  1073--1083.

\bibitem[{Sutskever et~al.(2014)Sutskever, Vinyals, and
  Le}]{sutskever2014sequence}
Ilya Sutskever, Oriol Vinyals, and Quoc~V Le. 2014.
\newblock Sequence to sequence learning with neural networks.
\newblock In \emph{Advances in neural information processing systems}, pages
  3104--3112.

\bibitem[{Wang et~al.(2019)Wang, Gao, Huang, and Zhou}]{wang2019concept}
Wenbo Wang, Yang Gao, He-Yan Huang, and Yuxiang Zhou. 2019.
\newblock Concept pointer network for abstractive summarization.
\newblock In \emph{Proceedings of the 2019 Conference on Empirical Methods in
  Natural Language Processing and the 9th International Joint Conference on
  Natural Language Processing (EMNLP-IJCNLP)}, pages 3067--3076.

\bibitem[{Warstadt et~al.(2019)Warstadt, Singh, and
  Bowman}]{warstadt2019neural}
Alex Warstadt, Amanpreet Singh, and Samuel~R Bowman. 2019.
\newblock Neural network acceptability judgments.
\newblock \emph{Transactions of the Association for Computational Linguistics},
  7:625--641.

\bibitem[{West et~al.(2019)West, Holtzman, Buys, and Choi}]{west2019bottlesum}
Peter West, Ari Holtzman, Jan Buys, and Yejin Choi. 2019.
\newblock Bottlesum: Unsupervised and self-supervised sentence summarization
  using the information bottleneck principle.
\newblock In \emph{Proceedings of the 2019 Conference on Empirical Methods in
  Natural Language Processing and the 9th International Joint Conference on
  Natural Language Processing (EMNLP-IJCNLP)}, pages 3743--3752.

\bibitem[{Zhang et~al.(2019{\natexlab{a}})Zhang, Zhao, Saleh, and
  Liu}]{zhang2019pegasus}
Jingqing Zhang, Yao Zhao, Mohammad Saleh, and Peter~J Liu. 2019{\natexlab{a}}.
\newblock Pegasus: Pre-training with extracted gap-sentences for abstractive
  summarization.
\newblock \emph{arXiv preprint arXiv:1912.08777}.

\bibitem[{Zhang et~al.(2019{\natexlab{b}})Zhang, Merck, Tsai, Manning, and
  Langlotz}]{zhang2019optimizing}
Yuhao Zhang, Derek Merck, Emily~Bao Tsai, Christopher~D Manning, and Curtis~P
  Langlotz. 2019{\natexlab{b}}.
\newblock Optimizing the factual correctness of a summary: A study of
  summarizing radiology reports.
\newblock \emph{arXiv preprint arXiv:1911.02541}.

\end{thebibliography}
